# A MAP-MRF filter for phase-sensitive coil combination in autocalibrating partially parallel susceptibility weighted MRI

Sreekanth Madhusoodhanan, Joseph Suresh Paul

*Abstract*— A statistical approach for combination of channel phases is developed for optimizing the Contrast-to-Noise Ratio (CNR) in Susceptibility Weighted Images (SWI) acquired using autocalibrating partially parallel techniques. The unwrapped phase images of each coil are filtered using local random field based probabilistic weights, derived using energy functions representative of noisy sensitivity and tissue information pertaining to venous structure in the individual channel phase images. The channel energy functions are obtained as functions of local image intensities, first or second order clique phase difference and a *threshold scaling parameter* dependent on the input noise level. Whereas the expectation of the individual energy functions with respect to the noise distribution in clique phase differences is to be maximized for optimal filtering, the expectation of tissue energy function decreases and noise energy function increases with increase in threshold scale parameter. The optimum scaling parameter is shown to occur at the point where expectations of both energy functions contribute to the largest possible extent. It is shown that implementation of the filter in the same lines as that of Iterated Conditional Modes (ICM) algorithm provides structural enhancement in the coil combined phase, with reduced noise amplification. Application to simulated and in vivo multi-channel SWI shows that CNR of combined phase obtained using MAP-MRF filter is higher as compared to that of coil combination using weighted average.

*Index Terms*— Clique phase difference, CNR, Coil Combination, Markov Random Field, SWI.

## I. INTRODUCTION

The feasibility of accelerating Susceptibility Weighted Imaging (SWI) acquisitions by using parallel imaging techniques has shown that Generalized Autocalibrating Partially Parallel Acquisition (GRAPPA) has advantage over SENSitivity Encoding (SENSE) in terms of its ability to handle motion induced artifacts and field inhomogeneities [1-3]. However, GRAPPA is a channel-wise reconstruction technique, leading to spatially varying amplification of the receiver noise. The resulting influence of noise in channel phase introduces signal losses (loss of venous structural information) and reduced Signal-to-Noise ratio (SNR) during the process of coil combination following GRAPPA reconstruction. The signal losses also occur in the form of cusp artifacts [4] due to improper phase combination. A combined measure of the signal losses and noise amplification using a modified form of the Contrast-to-Noise Ratio (CNR), is therefore, more appropriate for assessment of SWI processing. Same values of CNR in two cases may indicate less losses and higher noise level, or higher losses and low noise level. For discrimination of the above two conditions, a noise dependent filtering parameter further referred to as the "*threshold scaling parameter*" is to be determined using a prior calibration procedure. Since the observed channel phase is the sum of a tissue phase and sensitivity induced noisy phase component, maximizing the CNR requires a combination method based on prior filtering of the channel phases using an optimum value of the threshold scaling parameter.

Enhancement of venous structures is achieved by means of a phase mask obtained using linear mapping of the high-pass filtered phase (Haacke et al., 2009; Mittal et al., 2009) [5-6]. Other type of non-linear phase mask [7] has also been proposed to improve the CNR by minimizing filter induced noise amplification. Phase masks may accentuate either negative or positive phase effects and made up of values between 0 and 1, with most values falling around 1. In the statistical distribution of the high-pass filtered phase, information about vascular structures are contained mostly in the tails of the distribution. Depending on the dipole orientation of the individual structural element, one of the tail portions emphasizes the structure, while the other contains information relating to the peripheral (edge) region. This results in a more or less symmetric form of distribution of the high-pass filtered phase. It is to be emphasized that the distribution of high-pass filtered phase is similar in characteristics to those of first or second order clique phase differences computed at each voxel of an unwrapped phase image. The central part of the symmetric distribution is representative of the noisy component, including background effects. Since the aforementioned characteristics are also applicable to the individual channel phases, contributions of noise and structural phase components can be estimated using a threshold ($\pm tr_j$) separating the central region from its tails. Since the shape of channel phase distributions are inherently complex in nature, direct application of statistical learning algorithms to obtain either MAP [8], or ML [9-10] estimates of the filter weights is extremely cumbersome.

The priors computed using the observed distribution and global threshold are, however, not sufficient to model the contextual dependence of noise at locations corresponding to varying magnitude levels in the individual channel phase images. Based on the local channel intensities, this essentially points to using information about the local phase variance for characterising the phase interactions among neighboring voxels (cliques). These are modelled using appropriate forms of energy functions representative of the tissue and noise induced phase components. Due to the fact that venous information is contained in the tail regions of the clique phase difference distribution, energy function representing the tissue phase must possess large values for phase differences exceeding the local threshold. Estimation of the local threshold is premised on the fact that the extent of noise induced phase variations is inversely related to the local signal intensity, indicative of the SNR at each voxel.

In view of the local contextual dependence to model the contribution of noise and tissue phase components, we propose to use a Markov Random Field (MRF) [11] to estimate the individual contributions. Due to the presence of noise, the first or second order clique phase differences are treated to be random variables. MRF priors signifying contributions of noise and tissue induced phase variation are derived using local thresholds and appropriate forms of energy functions. The interplay of energy function models and local thresholds are such that with increase in threshold values, the expectation of tissue energy function decreases and that of noise energy function increases. Ideally, the optimal selection is made at the point where both are maximized at the same time. This ensures that thresholds larger than the optimum value result in better noise reduction but loss of venous structures and vice-versa.

The proposed method is detailed in Section II. Determination of the effects of noise on choice of optimum thresholds is presented in the simulation studies part of Section III (A)-(B). Results using in vivo SWI data are included in the experimental studies part of Section III (C). Discussion and summary is provided in Section IV.

## II. THEORY
### A. Pre-processing of multi-channel SWI phase data

The channel phase consists of components proportional to the product of field variation and echo time, in addition to that contributed by noise. The field variation in each coil is a combination of the background field and the local field. The influence of large background field induces cusp artifacts and signal losses in the combined phase image. A second problem is due to the presence of a local background field gradient, leading to a shift of the echo center in $k$-space and linear phase shift [12]. The linear phase shift is removed by detecting the echo peak and shifting the $k$-space data such that the signal center occurs at the actual $k$-space center.

Usually, the background phase has slowly varying spatial characteristics and a suitable high-pass filter is used to remove the background phase variations. In its most basic form, the high-pass filtering operation is performed by subtracting averaged phase values of neighboring voxels from the unwrapped phase. The neighborhood size used for effective filtering of background phase typically varies with distance from the brain center [4], especially near to the air-tissue interface. However, since we do not use the neighborhood operations for a direct high-pass filtering of the channel phase, a fixed neighborhood system is used for deriving the MRF weights. The sample neighborhood system is shown in Fig. 1.

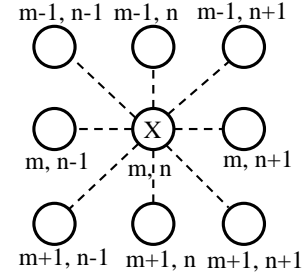

Fig.1:- Second order neighboring system with the center of 3 × 3 mask is labelled as X.

Denoting the unwrapped phase by $\varphi_j(m,n)$, the high-pass filtered phase

$$\varphi_{jF}(\bar{r}) = \varphi_{jF}(m,n)$$
$$= \varphi_j(m,n) - \frac{1}{N}\sum_{p,q=-1}^{1}\varphi_j(m+p,n+q) \quad (1)$$

A scaled version of $\varphi_{jF}$ can be equivalently represented using a central difference computed at each clique (clique phase difference).

$$\varphi_{jF}(m,n) = N\varphi_j(m,n) - \sum_{p,q=-1}^{1}\varphi_j(m+p,n+q) \quad (2)$$

$\varphi_{jF}$ mainly consists of two components viz., a noisy phase component $\vartheta_j$, and a tissue related component $\theta_j$. With the nature of smooth variation of background phase, its effects are assumed to be contained in $\vartheta_j$. However, cusp artifacts arising from phase wraps are more sensitive to high-pass filtering as the relative noise levels in the channel phase being higher in regions with low intensity. Hence a direct application of high-pass filtering will intensify the effect of cusp artifacts in the combined phase. This can result in loss of venous information during coil combination. In the proposed method, we bypass the high-pass filtering of channel phases using a statistical filter that serves to reduce the noise and enhance venous information prior to coil combination.

### B. Statistical approach for phase sensitive coil combination
#### B1. Channel phase model
Let $S = \{S_i | i \in Z_N\}$ where $Z_N = \{1,2,\cdots,N\}$ represent a regular lattice structure with $N$ sites. Let $\Theta = \{\theta_i | i \in Z_N\}$ be the tissue phase and $\Psi_j = \{\varphi_{ij} | i \in Z_N\}$ be the measured phase in the $j$-th channel at each site. For ease of notation, the collection of clique phase differences from all sites $S = \{S_i | i \in Z_N\}$ is denoted by $\Delta\varphi_j$. The phase differences are used only to estimate the clique potentials and energy functions. We describe the overall pdf of $\Delta\varphi_j$ as a weighted combination of components contributed by the tissue and noise induced phase variations.

$$f(\Delta\varphi_j) = w_{tj}f_t(\Delta\varphi_j) + w_{cj}f_c(\Delta\varphi_j), \quad w_{tj} + w_{cj} = 1. \quad (3)$$

with the notion that tissue phase variations are abrupt, it is justifiable to consider $f_t(\Delta\varphi_j = 0) = 0$ and hence $w_{cj} = \frac{f(0)}{f_c(0)}$. In sequel, we obtain

$$f_t(\Delta\varphi_j) = \frac{f(\Delta\varphi_j) - w_{cj}f_c(\Delta\varphi_j)}{(1 - w_{cj})} \quad (4)$$

A best fitting model for $f_c(\Delta\varphi_j)$ is assumed to follow a distribution similar to the form of marginal distribution of phase [13] when the real and imaginary components possess zero mean normal distribution with a prior known variance. For a given input SNR, the distribution takes the form

$$f_c(\Delta\varphi_j) = \exp\left(-\frac{\alpha^2}{2\pi}\right)\left[1 + \sqrt{\pi}\beta \exp(\beta^2)(1 + \text{erf}(\beta))\right] \quad (5)$$

where $\alpha = \frac{SNR}{\sqrt{2}}$ and $\beta = \alpha \cos(\Delta\varphi_j)$. A best fitting model as a function of the input SNR is obtained by minimizing the chi-square error

$$\chi^2 = \sum_i \frac{\left[f\left(\Delta\varphi_{j_i}\right) - f_c\left(\Delta\varphi_{j_i}\right)\right]^2}{f_c\left(\Delta\varphi_{j_i}\right)} \quad (6)$$

where $i$ denotes the index of observations in the measured distribution and samples in the model density function. With the peak intensity normalized to unity, the noise standard deviation is measured as σ=1/*SNR* at the minimum point. Fig. 2(a) illustrates the chi-square error plot as a function of input SNR and the sample distributions (solid line) along with the estimated model (dotted line) for three different input SNRs. The best possible fit is obtained when the chi-square error is minimum.

The same procedure applied to the combined phase yields a threshold $\pm 1/(\sqrt{2}\,\alpha)$, providing a measure of the noise level. Thus the area enclosed between the observed (high-pass filtered phase) and noise distribution (model) provides a measure of tissue information retained in the combined phase. Since this area reduces with increase in noise, or equivalently the threshold, it is an appropriate representation of the CNR. This is illustrated in Fig. 2(b).

*B2. MAP-MRF Framework for weight estimation*

MAP estimation maximizes the posterior probability $p(\Theta|\Psi_j)$. The posterior probability can be estimated according to Bayes theorem, $p(\Theta|\Psi_j) \propto p(\Theta)p(\Psi_j|\Theta)$ where $p(\Theta)$ is evaluated using the local relationships between sites in each channel and $p(\Psi_j|\Theta)$ is estimated using a model fitting of the distribution of $\Delta\varphi_j$ as described in *Section* II-*B*1. The MRF model is used

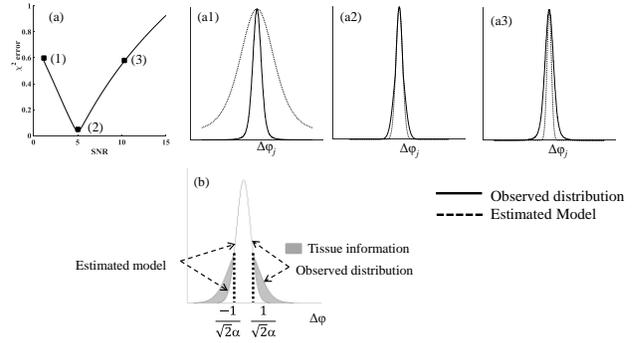

Fig 2:-(a) $\chi^2$ error as a function of SNR for individual channel images. (aX) Distributions of observed (solid line) channel clique phase difference and the estimated (dotted line) model for three input SNRs (1)-(3). (b) Distribution of observed clique phase difference and the model estimated from combined phase. The observed distribution deviates from the model at $\pm 1/\alpha\sqrt{2}$. The shaded area provides a measure of CNR.

to obtain the MAP estimation in an efficient manner as reported in other similar applications [14]. We assume that the local relationships between sites in $S$ are described by a neighbourhood system $\mathfrak{N} = \{\mathfrak{N}_i | i \in Z_N\}$, where $\mathfrak{N}_i \subseteq S$ representing a set of sites adjacent to the site $S_i$ as in Fig. 1. According to the Hammersley-Clifford theorem, $\Theta$ is an MRF with respect to $\mathfrak{N}$ if and only if $p(\Theta)$ is a Gibbs distribution with respect to $\mathfrak{N}$. A Gibbs distribution of $\Theta$ is given by

$$p(\Theta) = \frac{1}{Z_j}\exp\left[\frac{-U_{tj}(\Theta)}{T}\right] \quad (7)$$

where $T$ is a temperature parameter, $U_{tj}(\Theta)$ is the potential function representative of tissue phase interactions, and $Z_j$ is a normalizing constant given by

$$Z_j = \exp\left[\frac{-U_{tj}(\Theta)}{T}\right] + \exp\left[\frac{-U_{cj}(\Theta)}{T}\right] \quad (8)$$

$U_{cj}(\Theta)$ is the potential function representative of the noise-induced phase variation. From (7) and (8), the MAP decision rule becomes

$$\widehat{\Theta} = \underset{\Theta}{argmax}\frac{1}{Z'_j}\exp\left[\frac{-\left(U_{tj}(\Theta) + U_{tj}(\Psi_j|\Theta)\right)}{T}\right] \quad (9)$$

where

$$Z'_j = \exp\left[\frac{-\left(U_{tj}(\Theta) + U_{tj}(\Psi_j|\Theta)\right)}{T}\right] + \exp\left[\frac{-\left(U_{cj}(\Theta) + U_{cj}(\Psi_j|\Theta)\right)}{T}\right] \quad (10)$$

$U_{Xj}(\Theta)$ and $U_{Xj}(\Psi_j|\Theta)$ denote the respective prior and likelihood potentials characterizing the tissue and noise related interactions.

Energy functions are used to characterize each type of interaction at a clique such that the overall potential at a given location (*m,n*) is determined by summing up the values of energy functions at each clique. We follow the notations used in *Section* II-*A* to describe the clique interactions. For a given type of interaction $X \epsilon [t, c]$, the potential function $U_X$ is defined as

$$U_X(S_i) \triangleq U_X(m,n)$$
$$= \sum_{p,q=-1}^{1} \Phi_X(\varphi(m,n) - \varphi(m+p, n+q)) \quad (11)$$

where $\Phi_X$ denotes the energy function signifying the particular type of interaction, expressed as a function of the phase difference at each clique. For ease of notation, the collection of clique phase differences from all sites $S = \{S_i | i \in Z_n\}$ is denoted by $\Delta\varphi$.

The derived objective function in (9) can be optimized by the Iterated Conditional Mode (ICM) method [15]. ICM persistently seeks a lower energy configuration and never allows increase in energy, which guarantees a faster convergence rate. ICM assumes that 1) the observed variable $\Psi_j = \{\varphi_{ij} | i \in Z_N\}$ are conditionally independent and 2) The state of $S_i$ depends only on the state of its adjacent sites $\mathfrak{N}_i$ (Markovian property). These two assumptions allow the minimization of its local energy terms

$$\sum_{p,q=-1}^{1} \Phi_X\left(\varphi_{ij}(m,n) - \varphi_{ij}(m+p, n+q)\right) + U_X(\varphi_{ij}|\theta_i) \quad (12)$$

where

$$U_X(\varphi_{ij}|\theta_i) = \sum_{p,q=-1}^{1} f_X\left(\varphi_{ij}(m,n) - \varphi_{ij}(m+p, n+q)\right)$$

*B3. Algorithm for channel phase filtering*

The local thresholds characterize the influence of noise on the channel phase. As increase in noise influences $\Delta\varphi_j$ in regions of low sensitivity, the local threshold for estimation of priors in (9) is determined using

$$tr_j(\bar{r}) = \frac{c_j}{|I_j(\bar{r})|} \quad (13)$$

The constant $C_j$ is related to the maximum channel intensity using a threshold scaling parameter $K$ such that $C_j = KI_{jmax}$. In later sections, it is shown that optimal $K$ and hence $tr_j$ is dependent on the input noise level. As shown in subsequent section, reducing $K$ below the optimal value leads to increase in noise in the combined phase but with reduced loss of tissue information. Likewise, increasing $K$ to values more than the optimum is accompanied by increase in loss of tissue information and reduced noise. This means that two points on either side of the optimum are representative of combined phase having the same value of CNR. Of these, one corresponds to an output with more tissue information accompanied by higher noise level, and the other with loss of tissue information accompanied by reduced noise level. It is to be noted that although the clique phase difference distribution of the combined and channel phases look similar, its shape is dependent on the chosen value of $K$. That is, choice of a lower $K$ than the optimum will result in a lower value for $\alpha$ and vice-versa in the model (4) when applied to the combined phase image.

Using an appropriate choice of $K$, the ICM algorithm is employed to maximize the probability of tissue phase $p(\theta_i|\varphi_{ij})$. The proposed algorithm for phase-sensitive coil combination is summarized below. It is assumed that the threshold scaling parameter $K$ is already determined using a prior calibration.

---

**Algorithm 1**: ICM channel phase filtering

---
*Iterate for $k = 1, \cdots, MaxIter$:*
  *Iterate for $i = 1, \cdots, N$:*
    *Iterate for $j = 1, \cdots, n_c$ :*
      1) *Compute* $W_{tj}^{(k)}, f_{tj}^{(k)}, W_{cj}^{(k)}, f_{cj}^{(k)}$
      2) *Compute* $U_{tj}^{(k)}(\theta_i) + U_{tj}^{(k)}(\varphi_{ij}|\theta_i)$ *and* $U_{cj}^{(k)}(\theta_i)$
        $+ U_{cj}^{(k)}(\varphi_{ij}|\theta_i)$
      3) *Compute* $p^{(k)}(\theta_i|\varphi_{ij})$
        $= \frac{1}{Z} exp\left[\frac{U_{tj}^{(k)}(\theta_i) + U_{tj}^{(k)}(\varphi_{ij}|\theta_i)}{T}\right]$
      4) *set* $\varphi_j^{(k+1)} = \varphi_j^{(k)} p^{(k)}(\theta_i|\varphi_{ij})$
      5) *repeat step* $1 \cdots 4$ *until* $\|\varphi_j^{(k+1)} - \varphi_j^{(k)}\|$
        $< tol,$ *or* $k = MaxIter.$
    *end*
  *end*
*end*
$Ires = \sum_j |I_j| \exp(\varphi_j^{(k)})$

---

*B4. Determination of Energy Functions*

In the absence of local phase variation contributed by tissue and ideal case of *zero*-noise, the distribution of $\Delta\varphi$ may be approximated using an impulse function. The energy functions signifying noisy phase contribution should have the characteristics that the expectation of the function with respect to the noise distribution increases with increase in noise variance. In the noise-free case, the shape of energy function characterizing clique interactions of the noise induced phase variation should be such that the threshold scaling parameter must ideally tend to zero in order to maximize energy function. Due to the impulse shape, it is also required that the function has a maximum at $\Delta\varphi = 0$. This requires $K = 0$ for maximizing $\mathbb{E}[U_c] = \sum_j \mathbb{E}[U_{cj}\delta(\Delta\varphi_j)]$. A particular form of the function that makes the shape an impulse as $K$ tends to zero is

$$\Phi_{cj} = \frac{1}{1 + \frac{1}{K^2}\left\|\frac{|I_j|\Delta\varphi_j}{|I_{jmax}|}\right\|^2} \quad (14)$$

The energy function signifying the tissue contribution should have the characteristics that its expectation is maximized with increasing noise variance. This is true when the phase differences at each clique assume values higher than $\pm tr_j$. Further, the polarity of the phase mask determines the sign of phase values representative of venous structure and its immediate periphery. For example, in conventional SWI processing, usage of a positive phase mask enhances venous structures having positive values of high-pass filtered phase in the intravenous region and vice-versa in the immediate periphery (edges). In the case of high-pass filter using (2)-(3) with second-order neighbourhood, this is equivalent to the intravenous regions mostly possessing large positive values and peripheral regions (edges) exhibiting large negative values. Further, the histogram of high-pass filtered phase exhibits a symmetric shape; similar to that constructed using clique phase differences. Thus it is necessary to model the phase contributions of venous structures together with the edge region using a symmetric form of energy function having high values for $|\Delta\varphi_j| \geq tr_j$. For a positive phase mask in particular, the components of energy function signifying intravenous structures and surrounding edges are maximized when the high-pass filtered phase values lie above $+tr_j$, and below $-tr_j$ respectively. Maximizing the energy function components in both of the above conditions require the forms of respective tissue related energy functions to consist of a pair of switching functions shifted by $tr_j$ in both positive and negative directions. These can be implemented either using sigmoid functions, or error functions. The respective forms of energy functions are shown below.

| Sample energy function | |
|---|---|
| $\Phi_{tj} = \begin{cases} \frac{1}{1+\exp\left(\frac{\Delta\varphi_j}{tr_j}+1\right)}, & \text{for } \Delta\varphi_j \leq -tr_j \\ \frac{1}{1+\exp\left(-\frac{\Delta\varphi_j}{tr_j}+1\right)}, & \text{for } \Delta\varphi_j \geq +tr_j \end{cases}$ (15a) | |
| $\Phi_{tj} = \begin{cases} \frac{1-\text{erf}(\Delta\varphi_j+tr_j)}{2}, & \text{for } \Delta\varphi_j \leq -tr_j \\ \frac{1+\text{erf}(\Delta\varphi_j-tr_j)}{2}, & \text{for } \Delta\varphi_j \geq +tr_j \end{cases}$ (15b) | |

*B5. Effect of noise*
With presence of noise, the phase term in both energy functions is treated as a random variable. Consequently, the scaling parameter $K$ can be optimized only by maximizing the expectation $\mathbb{E}[\Phi_{cj}]$

$$\mathbb{E}[\Phi_{cj}] = \frac{1}{N}\sum_{\bar{r}}\int_{-\infty}^{\infty}\frac{f_c(\Delta\varphi_j)}{1+\left(\frac{\Delta\varphi_j}{tr_j}\right)^2}d\Delta\varphi_j \quad (16)$$

Here, $\Delta\varphi_j$ denotes the phase difference at each clique and $f_c(\cdot)$ denotes the distribution of $\Delta\varphi_j$ due to noise. Assuming the noise to be Rician, the non-linear mapping due to $\tan^{-1}(\cdot)$ operation makes the shape of distribution $f_c(\cdot)$ to be of complex nature [13]. However, at regions of high SNR, it can be approximated by a Gaussian

$$f_c(\Delta\varphi_j) = \frac{1}{\sqrt{2\pi(\sigma/|I_j|)^2}}\exp\left[\frac{-\Delta\varphi_j^2}{2(\sigma/|I_j|)^2}\right] \quad (17)$$

Using the above form for $f_c(\cdot)$,

$$[\Phi_{cj}] = \frac{\pi K|I_{jmax}|}{\sqrt{2\pi\sigma^2}}\exp\left[\frac{K^2|I_{jmax}|^2}{2\sigma^2}\right]\text{erf}\left[\frac{K|I_{jmax}|}{\sqrt{2}\sigma}\right] \quad (18)$$

Numerical simulation of (18) shows that for a given input noise level ($\sigma$), $\mathbb{E}[\Phi_{cj}]$ increases with increasing $K$ values. In similar lines to (18), the expectation with respect to $f_c(\cdot)$ of the error function representation for tissue energy function yields

$$\mathbb{E}[\Phi_{tj}] = \frac{1}{N}\sum_{\bar{r}}\int_{-\infty}^{\infty}\frac{1}{\sqrt{2\pi(\sigma/|I_j|)^2}}\exp\left[\frac{-\Delta\varphi_j^2}{2(\sigma/|I_j|)^2}\right]$$

$$\left\{\left[\frac{1-\text{erf}(\Delta\varphi_j+tr_j)}{2}\right] + \left[\frac{1+\text{erf}(\Delta\varphi_j-tr_j)}{2}\right]\right\}d\Delta\varphi_j$$

$$\frac{1}{N}\sum_{\bar{r}}\frac{1}{\sqrt{2\pi(\sigma/|I_j|)^2}} - \text{erf}\left[\frac{K|I_{jmax}|}{\sqrt{|I_j|^2+2\sigma^2}}\right] \quad (19)$$

Fig. 3 (a1)-(a3) show the expectations as a function of $K$ for input noise levels $\sigma = 0.003$, 0.007 and 0.011. It is observed that for each case, $\mathbb{E}[\Phi_c]$ increases and $\mathbb{E}[\Phi_t]$ decreases with increase in $K$. Meeting point of the two expectations correspond to the best possible choice of $K$. Increase in $K$ is expected to result in noise reduction due to increase in $\mathbb{E}[\Phi_{cj}]$, and loss of structural information due to decrease in $\mathbb{E}[\Phi_{tj}]$.

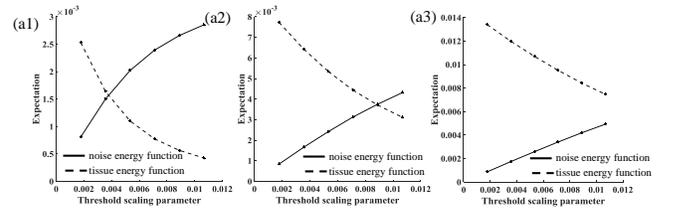

Fig 3:- (a1)-(a3) Theoretical expectations of energy functions as function of the threshold scaling parameter shown for three different noise levels. (a1) $\sigma = 0.003$, (a2) $\sigma = 0.007$ and (a3) $\sigma = 0.011$

## III. RESULTS

*A. Generation of synthetic multi-channel SWI for determination of effect of noise on threshold scaling parameter*

This section describes the Monte carlo approach to numerically verify the effects of noise predicted by the analysis in section II.*B*. Multi-coil data are synthesized by multiplication of a reference image (a single coil SWI image) with complex coil sensitivities simulated using Biot-Savarts law. It is to be emphasized that the noise inherently present in the single coil image will be unaffected by the filtering procedure. Hence the performance of the proposed filter can only be compared against the unfiltered form of coil combination. The individual channel images are first scaled by division with the maximum intensity across all channels. Complex Gaussian noise samples of known variance are then added to the scaled channel images. For each channel, the clique phase difference distribution is fitted with the model in (6) using minimum chi-square error criterion. The maximum value of $tr_j$ is limited to $1/\sqrt{2}\alpha_j$ computed at the time of model fitting. This prevents outliers due to low intensity values.

*B. Practical approach for choice of K*

In practice, the $\sigma$ values are not known. Therefore, SWI coil combination should be preceded by a calibration phase in which an ad hoc search is performed to select the best choice of *K*. In this approach, we start with a small value of *K* to first estimate the local thresholds and MRF priors. The unwrapped channel phases are weighted by the MRF priors before coil combination. Histogram of the combined high-pass filtered phase is fitted with (6). The CNR measured as the area within the histogram and lying outside the region enclosed by the model is plotted as a function of *K* in Fig. 4. Panels (a1)-(a3) show CNR plots obtained for three input noise levels with standard deviations of 0.003, 0.007, and 0.011. Fig. 5 (a)-(c) depict the changes in enclosed area for the three noise levels. The area indicative of CNR is seen to decrease with increase in noise level. Fig. 4 (a1)-(a3) also demonstrates that the CNR increases with *K* until it reaches a peak value. In this phase of the plot, the extent of reduction in tissue structure is less as compared to the amount of noise reduction. Following the peak, the relative reduction in structure is more than that of noise with accompanying reduction in CNR.

Row-wise panels of Fig. 6 (a)-(b) show the combined images obtained for an input noise level of 0.003 without application of MAP-MRF filter. Of these, panels (a) illustrate the phase mask constructed from homodyne filtered phase, and (b) correspond to variable high-pass derived phase mask [4]. It is seen that both phase masks have signal losses due to cusp artifacts and noise. Bottom panels (c)-(d) show the combined images obtained by MAP-MRF filtering of the individual channel phases for a *K*-value of 0.004. It is clearly seen that signal losses are reduced in both cases. The rightmost panels display a portion of the magnitude SWI images enclosed by the rectangular bounding box. As compared to (a), CNR is higher in (c) due to reduced noise amplification. The same observation applies to (b) versus (d). The arrows point to regions with missing information retrieved using MAP-MRF filtering.

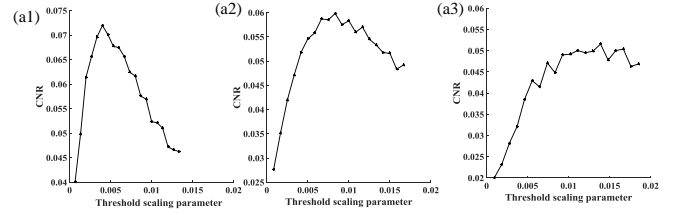

Fig 4:- (a1)-(a3) CNR measured from the combined phase image shown as function of the threshold scaling parameter shown for three different noise levels. (a1) $\sigma = 0.003$, (a2) $\sigma = 0.007$ and (a3) $\sigma = 0.011$. CNR is measured using the area enclosed between the combined phase difference histogram and the minimum chi-square fitted model.

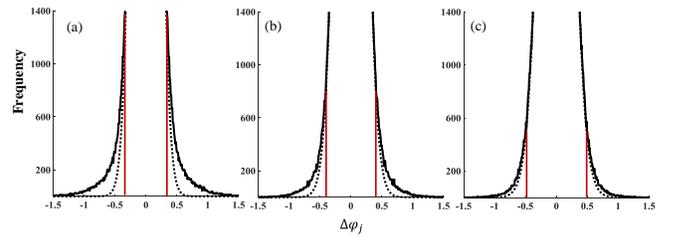

Fig. 5:- (a)-(c) Histogram of the combined high-pass filtered phase fitted with the model in Eq. (6) for channel input noise standard deviations of 0.003, 0.007and 0.011 respectively. The noise is added to simulated channel images after scaling the intensities by the maximum intensity across all channels. The dotted curve represents the fitted model. The area enclosed between the observed histogram and the model is seen to decrease with increase in input noise level. This area is an indicator of CNR in the combined phase image.

*C. Application to real data*

Raw k-space data were acquired from Siemens 1.5T Magnetom-Avanto clinical MR scanner at Sree Chitra Tirunal Institute of Medical Sciences and Technology, Trivandrum, India. All subjects were scanned with prior written informed consent as recommended by the institutional ethics committee. Three datasets were acquired using sixteen-channel head array coils using 3D-GRE SWI sequence and acceleration factor *R*=2 (TE = 30 ms, TR = 49 ms, slice thickness = 2.0 mm, FOV = 240 mm and matrix size of 480 × 480). After GRAPPA reconstruction and channel phase correction, the channel phases are unwrapped prior to the calibration for determination of optimal *K*. Fig. 7 shows the calibration plots for each dataset along with respective phase masks obtained after combination with MAP-MRF filtering of channel phases. The optimum *K* values for each dataset are found to be 0.0012, 0.0015 and 0.0008. It is seen that noise levels are higher when the combination is performed using scale parameter value less than the optimum. Loss of venous structures and reduced noise levels are observed in the phase mask obtained using scale parameter larger than the optimum value. The combination using optimum *K* (corresponding to the peak in the calibration curve) yields a better quality image without significant structural losses and noise amplification. The rightmost panels display a portion of the magnitude SWI images enclosed by the rectangular bounding boxes.

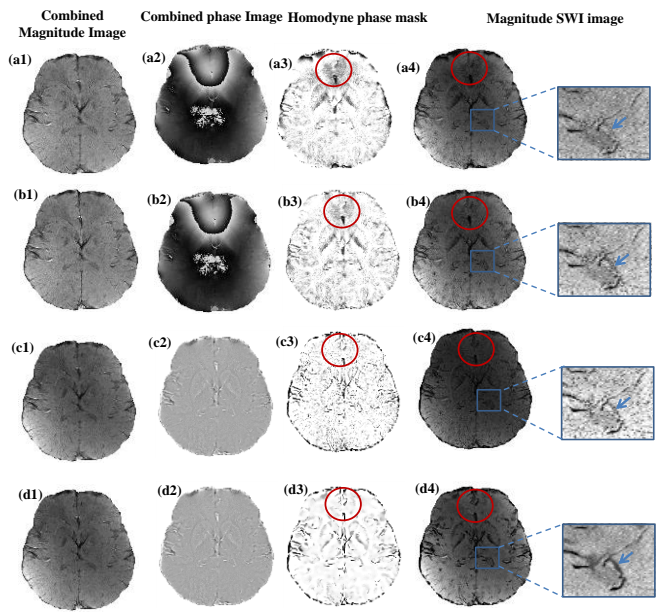

Fig.6:- (X1)-(X4)) Combined magnitude image, phase image, phase mask, and magnitude SWI image. Red circles are used to compare region with signal losses. (a) homodyne phase mask obtained from combined phase image without application of the MAP-MRF filter to individual channels. (b) variable high pass filtered phase mask obtained from combined phase image without application of the MAP-MRF filter to individual channels. (c) homodyne phase mask obtained from combined phase image with application of the MAP-MRF filter to individual channels. (d) variable high pass filtered phase mask obtained from combined phase image with application of the MAP-MRF filter to individual channels.. Blue bounding box in (c4)-(d4) is used to highlight the CNR improvement in combined image. Blue arrows point to region with missing information retrieved using MAP-MRF filter.

## IV. DISCUSSION AND SUMMARY

In this paper, we have introduced the use of MAP-MRF framework for modelling the channel phase contribution due to noise and intrinsic susceptibility variations. Experiments carried out on three clinical SWI images illustrated the applicability of the MRF model, and showed that the ICM algorithm can help in maximizing the posterior probabilities of true tissue phase. The combined phase image reconstructed after application of the MAP-MRF filter to each channel phase is found to be effective in minimizing signal losses due to noise and cusp artifacts.

One of the main contributions of this paper is the way of filtering out the effects of noise in each channel phase using thresholds derived from the clique phase difference distribution. The idea relating the presence of tissue related information in the tail portions of this symmetric form of distribution, and noise together with background effects in the central part is used for further estimation of the likelihood functions in the MAP-MRF framework. The mixture models are formulated using the clique phase differences. Due to the inverse tangent operation, the shape of phase distribution typically becomes complex and hence difficult to model. Consequently, a straightforward application of EM class of algorithms becomes difficult. For the present purpose, a minimum chi-square fit is found to be sufficient for tracking the noise distributions and estimation of CNR from the combined phase.

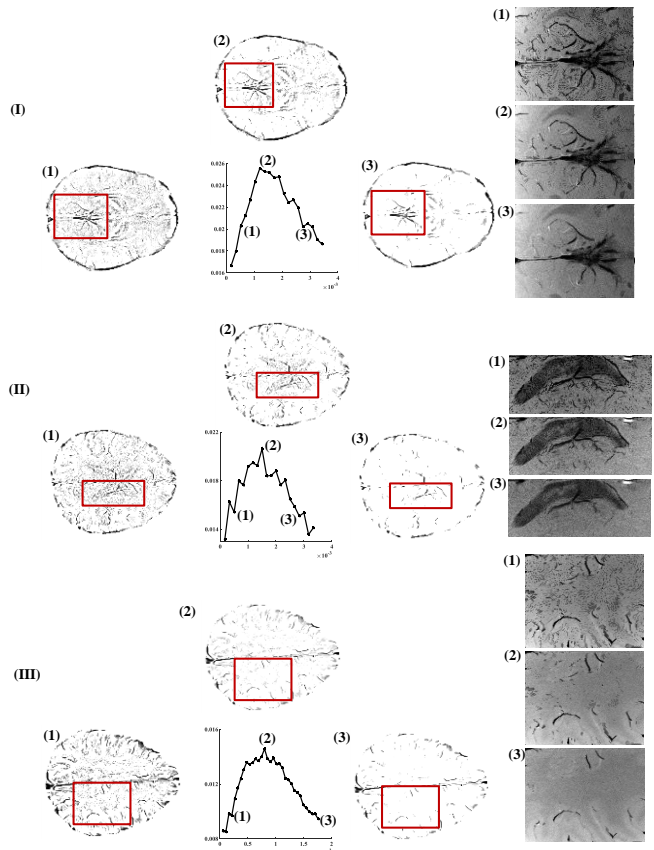

Fig.7:- Panels (I)-(III) show the combined phase masks for datasets I-III. The phase masks obtained using MAP-MRF filter is shown for three different $K$ values (1)-(3). Calibration curves for each dataset are shown in the middle. In each case, the peak corresponds to the optimum $K$ value. Right most panels indicate regions within the highlighted areas of magnitude SWI images.

Although the MRF model works satisfactorily in mitigating noise while preserving structural information for relatively low SNR in selected number of channels, we find that it is hard to maximize the MAP estimates and result in slower convergence of the algorithm. The effective improvement in CNR is also limited as seen in Fig. 8. With higher input noise, the initial attainable CNR values are low. Despite the slow convergence, the percentage improvement observed in the final CNR is higher even at higher input noise level.

Unlike other combination techniques which rely on the sensitivity information, we have made use of the available magnitude information to locally determine the statistical features of the channel phase. We have formally introduced the idea of a local threshold involving a global scaling parameter independent of the channel images into the filtering process. In other words, the histogram of $\Delta\varphi_j$ does not depend on $K$. However, the histogram, of combined phase $\Delta\varphi$ depends on the choice of *K*. Another difficulty arises due to dependence of local threshold on local intensity value. If the intensity values are low, this introduces additional noise effects into the estimation. To eliminate this problem, an upper limit is assigned to the local threshold value in each channel based on the $\alpha_j$ value obtained from the model fitted to the channel-wise clique phase difference histogram. If the upper limit is chosen below this value, there is a reduction of the CNR estimated from the combined phase. As stated above, the range of this CNR reduction depends on the input noise level.

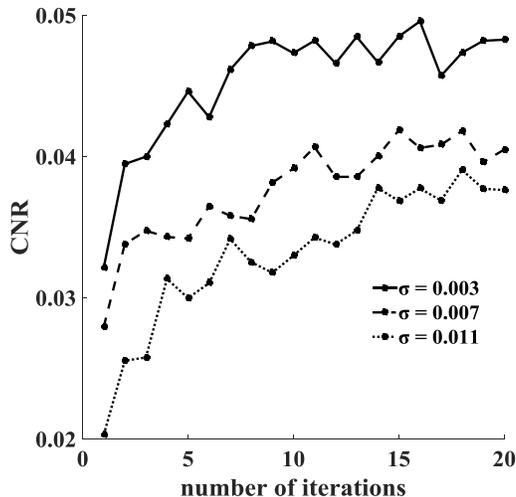

Fig. 8:- CNR improvement using ICM iterations.

Furthermore, the calibration phase requires tuning the *K*-value by repeatedly performing the model fitting. In this situation, the chi-squared model fitting is fast enough. An attractive feature of the fitting procedure is that the theoretical formulation using numerically computed expectations of the chosen forms of energy functions are found to be in close agreement with those obtained from measurement of CNR using histogram modelling of the combined phase. For each slice, the tuning needs to be performed separately. This prolongs the computation of mIP weighted images and forms a drawback of the proposed approach.